\documentclass[runningheads]{llncs}
\usepackage{graphicx}
\usepackage{amsmath,amssymb} 
\usepackage{color}
\usepackage{array}
\usepackage{hyperref}
\graphicspath{{figures/}}
\usepackage{xcolor}

\newcommand{\fig}[1]{Figure~\ref{fig:#1}}
\newcommand{\tab}[1]{Table~\ref{tab:#1}}

\usepackage{multirow}
\usepackage{wrapfig}

\def\argmax{\operatornamewithlimits{\rm arg\,max}}

\usepackage{graphicx}
\usepackage{amsmath}
\usepackage{amssymb}
\usepackage{booktabs}
\usepackage{comment}
\usepackage{booktabs}

\usepackage{xspace}

\newcommand\records{55 }

\begin{document}
\title{Hand-tremor frequency estimation in videos}

\titlerunning{Hand tremor frequency estimation in videos}
\authorrunning{S.L. Pintea, J. Zheng, X. Li, P.J.M. Bank, J.J. van Hilten, J.C. van Gemert}
\author{
    Silvia L. Pintea$^1$,
    Jian Zheng$^1$,
    Xilin Li$^1$,
    Paulina J.M. Bank$^2$,\break
    Jacobus J. van Hilten$^2$,
    Jan C. van Gemert$^1$
}
\institute{
    $^1$Vision Lab, Delft University of Technology, Netherlands\\
    $^2$Department of Neurology, Leiden University Medical Center, Netherlands
}

\maketitle
\begin{abstract}
We focus on the problem of estimating human hand-tremor frequency from input RGB video data. Estimating tremors from video is important for non-invasive monitoring, analyzing and diagnosing patients suffering from motor-disorders such as Parkinson's disease.
We consider two approaches for hand-tremor frequency estimation:
(a) a Lagrangian approach where we detect the hand at every frame in the video, and estimate the tremor frequency along the trajectory;
and (b) an Eulerian approach where we first localize the hand, we subsequently remove the large motion along the movement trajectory of the hand, 
and we use the video information over time encoded as intensity values or phase information to estimate the tremor frequency. 
We estimate hand tremors on a new human tremor dataset, \emph{TIM-Tremor}, containing 
static tasks as well as a multitude of more dynamic tasks, involving larger motion of the hands. 
The dataset has \records tremor patient recordings together with: 
associated ground truth accelerometer data from the most affected hand, RGB video data, and aligned depth data.
\end{abstract}

\begin{keywords}
Video hand-tremor analysis, phase-based tremor frequency detection, human tremor dataset, Eulerian hand tremors. 
\end{keywords}

\section{Introduction}
\label{sec:intro}
We focus on human hand-tremor frequency estimation from videos captured with common consumer RGB cameras.
The problem has a considerable importance in medical applications for aiding the medical personnel in the task of 
motor-disorder patient monitoring and tremor diagnosing \cite{di2017tremor,hssayeni2016automatic,ripin2017pathological,vidailhet2017simple}. 
Traditionally the clinical practice uses body-worn accelerometers which offer excellent measurements, yet is intrusive, slow to setup, 
and allows only measuring a single location per accelerometer. 
Replacing accelerometers with a common RGB camera brings forth a non-intrusive method of measuring full-body tremors, offering a strong advantage in the clinical practice.

In the context of tremor analysis, existing approaches require the use of specialized sensors \cite{dai2015quantitative,elble2016using,jeon2017automatic,kayaba2017non}, 
which makes it difficult to apply these methods in practice. 
Moreover, the targeted application of these approaches are the more high-level tremor diagnosing problem \cite{jeon2017automatic,roy2013learning} 
or tremor\slash no-tremor classification \cite{soran2012tremor}.
We propose to estimate human hand-tremor frequency from RGB videos, and compare against ground truth accelerometer data.

The main challenge, when performing human tremor frequency estimation, is the current lack of openly available realistic datasets. 
Existing work on human tremor analysis either evaluates using in-house data that is not publicly available \cite{dai2015quantitative,jeon2017automatic,xia2015new}, 
or on simulated tremor data where no ground truth tremor statistics are provided \cite{soran2012tremor}.
This limits the  assessment of human tremor analysis methods and, thus, its progress. An open evaluation dataset is needed. 

In this work: 
(i) we evaluate the frequency of human hand-tremors from RGB videos and we analyze two possible approaches: 
(i.a) a Lagrangian approach that focuses on the motion of the hand in the image plane, and estimates tremors over the hand positions;
(i.b) an Eulerian approach that aligns the hand position over a temporal window, by tracking it, and subsequently uses the image information over time as 
extracted from intensity values and phase-images, to perform a windowed Fourier analysis at every hand pixel;
(ii) we bring forth the \emph{TIM-Tremor} dataset, containing: \records RGB patient videos, 
together with associated ground-truth accelerometer recordings on the most affected hand, as well as aligned depth-data;
(iii) we analyze two variants of the Lagrangian approach and two variants of the Eulerian approach and evaluate them numerically on our proposed \emph{TIM-Tremor} human tremor dataset.

\section{Related work}
\label{sec:related_work}
\subsection{Motion analysis}
\noindent\textbf{Periodic motion.}
The work in \cite{polana1997detection} performs action recognition by using space-time repetitive motion templates.
Similar to using templates, in \cite{cutler2000robust} a self-similarity relying on time-frequency analysis is used for action recognition.
The work in \cite{goldenberg2005behavior} performs a spectral decomposition of moving objects to encode periodic motions for object recognition, 
while \cite{tralie2017quasi} performs eigen decomposition and describes periodic motion by the circularity or toroidality of an associated geometric space.
Following a similar trend, in \cite{lu2004repetitive} complex motion is decomposed into a sequence of simple linear dynamic models for motion categorization.
The work in \cite{ran2007pedestrian} focuses on pedestrian detection through periodic movement analysis. 
Similar to us, the work in \cite{soran2012tremor} performs tremor analysis, however in \cite{soran2012tremor} videos are classified into 
tremor \slash no-tremor using optical flow features and SVM.
In the recent work in \cite{2018park} a CNN is used for discriminating between Parkinson patients and non-Parkinson patients, using wrist-worn senors. 
In this work we also focus on periodic motion analysis, however our end goal is tremor frequency estimation rather than action recognition, object tracking or recognition.

Differently, in \cite{levy2015live,runia2018real,victor2017continuous} deep network architectures are trained for counting action repetitions. 
These actions must be clearly visible and recognizable in the camera view for the deep network architectures to work, 
while we focus on tremors which are subtle motions. 

The most similar work to our work is the work performed in \cite{uhrikova2010treman,uhrikova2011validation} where tremor frequency is measure from pixel intensities in the video.
However these methods assumes the location of the body part at which the tremor is measured to be known in advance and moreover, the frequency is estimated 
over intensity values rather than detected hand location over time, or image phase-information over time, as we propose here. 
The authors do not publicly provide either code or data, which makes it impossible for us to compare with their approach.  
 
\bigskip\noindent\textbf{Subtle motion.}
Small motion, difficult to see with the bare eyes, can be magnified~\cite{kooij2016depth,wadhwa2013phase} through a complex steerable pyramid. 
In the more realistic case, when the subtle motion is combined with a large motion, 
follow up work can magnify subtle motions such as tremors in the presence of large object motion such as walking~\cite{elgharib2015video,zhang2017video}. 
Video frequency analysis has been also employed for estimating the properties of physical materials~\cite{davis2015visual}. 
We also employ signal analysis in the Fourier domain, however rather than magnifying the subtle motion or estimating material properties, 
we estimate the frequency of the subtle tremor motion.
The works in \cite{jeon2017automatic,kayaba2017non} use specialized sensors or a digital light-processing projector, and a high frame-rate camera to detect small vibrations. 
Unlike \cite{jeon2017automatic,kayaba2017non}, we do not employ specific cameras or expensive sensors, we estimate the tremor frequency from common RGB videos. 

\subsection{Human body pose estimation}
Works such as \cite{bulat2016human,pishchulin2016deepcut} perform body pose estimation over multiple people, in deep networks.
In \cite{bulat2016human,carreira2016human,ramakrishna2014posemachines,wei2016convolutional} cascaded prediction or 
iterative optimizations are used for body pose estimation. 
We use the method in \cite{wei2016convolutional} for estimating where to measure the tremors. 
We opt for \cite{wei2016convolutional} due to its ease of usage and robustness. 
In this work we use the MPII Human Body Pose dataset \cite{andriluka20142d} for training the human body pose estimation models. 

\section{Hand-tremor frequency estimation}
\label{sec:method}
We start by localizing the affected hand. 
Subsequently, we consider two methods for hand-tremor frequency estimation: 
(a) Lagrangian hand-tremor frequency estimation, 
and (b) Eulerian hand-tremor frequency estimation. 

\subsection{Hand location estimation}
A first step in estimating human hand-tremors, is localizing the affected hand.
For this, we use the robust human body pose estimation proposed in \cite{wei2016convolutional}. 
This method provides us a hand location per frame $(x_i, y_i)$. 
We perform the tremor analysis on shorter temporal windows of the video, $w(t)$.

\begin{figure}[t]
	\centering
	\includegraphics[width=1.0\textwidth]{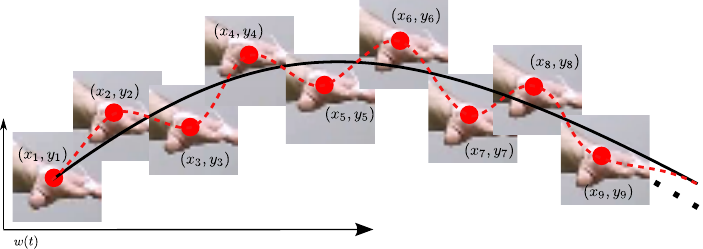}
	\caption{Lagrangian hand-tremor  estimation is based on frequency estimation of $(x,y)$ coordinates.
        We detect the hand position $(x_i, y_i)_{i \in w(t)}$ at every frame $i$ over a temporal window $w(t)$.
        The hand motion is characterized by a large  motion, depicted by the black line, and a small motion, depicted by the red dotted line.  
        We smooth this information over time, using a Kalman tracker to obtain the smooth coordinates of the hand. } 
	\label{fig:pose}
\end{figure}
\subsection{(a) Lagrangian hand-tremor frequency estimation}
\label{ssec:lag_method}
\fig{pose} depicts the idea behind the Lagrangian hand-tremor frequency estimation.
We start by detecting the hand locations $(x_i, y_i)_{i \in w(t)}$ over the temporal window $w(t)$.
The hand motion is typically characterized by a combination of two motions: a large hand trajectory motion, depicted through the continuous black line, 
and a small motion corresponding to the tremor, depicted in \fig{pose} by the dotted red line. 
We first apply a Kalman-filter tracker \cite{zarchan2013fundamentals} to the initial hand locations, detected by the pose estimation algorithm \cite{wei2016convolutional}.
This step is used for smoothing the hand trajectory, to obtain the large hand motion. 
We subsequently subtract this smooth trajectory from the original hand locations to retain only the \emph{x and y locations}  of the small hand motion, corresponding to the tremor.
Thereafter, we apply the windowed Fourier transform over these corrected locations. 
This provides us a PSD (Power Spectrum Density) function. 
We use the maximum frequency as the estimated hand-tremor frequency.

\begin{figure} 
	\centering
	\includegraphics[width=1.0\textwidth]{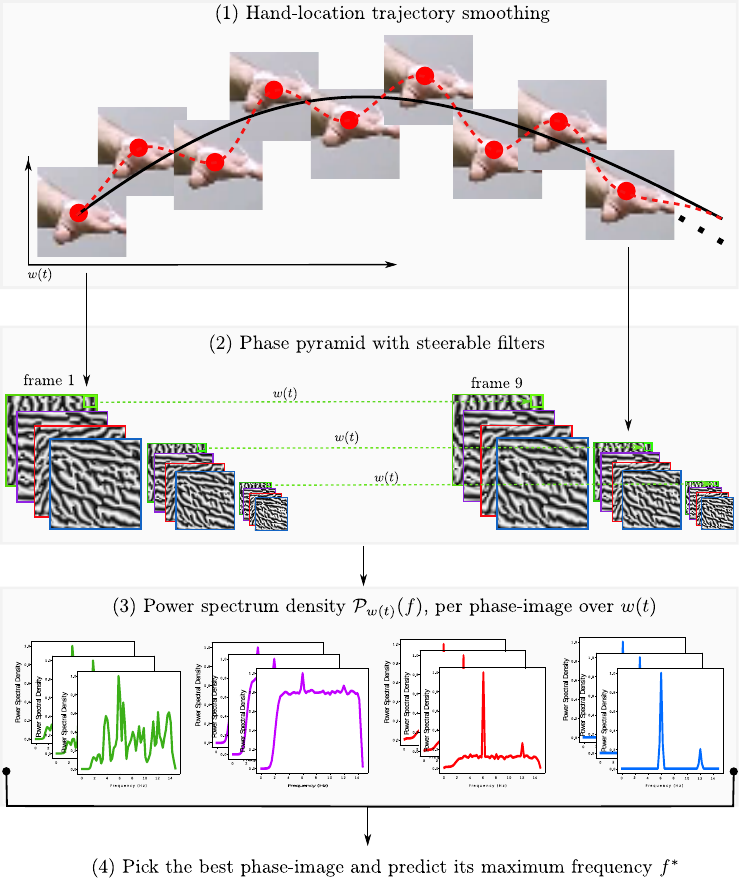}
	\caption{Eulerian hand-tremor estimation is based on frequency estimation in images.
        (1) The fist step is the same as in the Lagrangian illustrated in \fig{pose}: detecting a Kalman-filtered smoothed hand position  at every frame over a temporal window $w(t)$.
        (2) We crop image windows around the smoothed hand locations. 
        Each such cropped image is transformed into a phase-pyramid with 4 orientations and 3 scales using a steerbale filter bank. 
        (3) For every pixel, in every phase-image over the temporal window $w(t)$ we estimate a PSD (Power Spectrum Density).
        We accumulate these over the pixels in one phase-image, to obtain one PSD per phase-image. 
        (4) We select the most informative phase-image PSD and use it to estimate the tremor-frequency. 
    } 
	\label{fig:pose_eulerian}
\end{figure}

\subsection{(b) Eulerian hand-tremor frequency estimation}
\label{ssec:eul_method}
\fig{pose_eulerian} illustrates the Eulerian frequency estimation.  
The first step is the same as in \fig{pose}, where the hand locations are detected using the pose estimation method in \cite{wei2016convolutional}, 
and subsequently, we smooth the trajectory given by these hand detections using a Kalman tracker.
This gives us the smooth trajectory of the hand over time, in the video. 
We crop image windows around the temporally smoothed locations of the hand in the video --- along the black line depicted in \fig{pose_eulerian}.(1).
For each such image crop, we extract local motion information encoded as phase over different scales and orientations.
Thereafter, we compute the frequency of the hand-tremor by using the most informative phase-image.  
\fig{pose_eulerian} depicts these individual steps. 

\subsubsection{Phase-image computation.}
Works such as \cite{fleet1990computation,pintea2016making,wadhwa2013phase,zhang2017video} claim  that the phase responses over 
time contain descriptive information regarding the motion present in the image.  
In \cite{wadhwa2013phase} the use of complex steerable filters \cite{freeman1991design} is proposed for extracting local motion information. 
Given an input image $I(x,y)$ and a set of complex steerable filters of the form: $G_\sigma^\theta + i H_\sigma^\theta$, where $i=\sqrt{-1}$, $\sigma$ defines the scale of the filter, 
and $\theta$ the orientation, we obtain a complex steerable pyramid by convolving the image with this set of filters
\begin{equation}
( G_\sigma^\theta + i H_\sigma^\theta ) \circledast I(x,y) = A_\sigma^\theta(x,y) e^{ i \phi_\sigma^\theta(x,y)},
\end{equation}
where $\circledast$ denotes the convolution operations, and $A_\sigma^\theta(x,y)$ is the resulting amplitude for scale $\sigma$ and orientation $\theta$, 
and $\phi_\sigma^\theta(x,y)$ is the corresponding phase information. 
To obtain a phase-image, we set the amplitude to 1 and apply the inverse transformation \cite{freeman1991design} to reconstruct back the image. 
Examples of phase-images are depicted in \fig{pose}.(2).
We use 4 orientations: $\theta \in \{0, \frac{\pi}{4}, \frac{\pi}{2}, \frac{3 \pi}{4}\}$ and 3 scales: $\sigma \in \{1.0, 0.5, 0.25\}$, 
giving rise to 12 phase-images.
In addition to the 12 phase-images, we add the grayscale version of the cropped hand-image. 
Therefore, we have in total 13 images, which we merge into a single image with 13 channels, over which we estimate the hand-tremor frequency.  
\subsubsection{Hand-tremor frequency estimation.}
We filter each one of the 13 input channels over time with a 4$^{th}$-order Butterworth band-pass filter.
This eliminates noisy frequencies that cannot correspond to a natural human tremor.

To reduce the effect of the considered temporal window, $w(t)$, we use an adjustable Tukey window 
with the parameter $\alpha$ set to $\frac{f_s}{N - 1}$, where $f_s$ is the sampling rate and $N$ is the total number of frames in $w(t)$.
This ensures that the video signal over time is processed in a consistent manner while allowing for adjustable temporal window sizes, $w(t)$.

Within each temporal window, $w(t)$, we estimate a PSD function, over every input channel, at every pixel location.
For an input channel, $c$, we estimate the final PSD, $\mathcal{P}_{w(t)}^c (f)$, by averaging spatially the PSDs over the pixels in that channel.
We repeat this process for all 13 channels, giving rise to 13 PSD functions.

In \cite{cutler2000robust} the power spectrum is considered to be periodic at a certain frequency, $f$, if the PSD response at that frequency 
is at least a few standard deviations away from the mean PSD response.
This is indicative of how noisy is the PSD function.   
We use this same criterion to pick the most informative image channel; this is the channel over which we estimate the final hand-tremor frequency.
We define for each channel a score, $\mathcal{S}^c(f)$:
\begin{equation}
	\mathcal{S}^c(f) = \frac{1}{\mid w(t) \mid}\sum_{w(t)}\left( \mathcal{P}_{w(t)}^c (f) - \mu_{\mathcal{P}_{w(t)}^c} - k \sigma_{\mathcal{P}_{w(t)}^c} \right), 
    \label{eq:score}
\end{equation}
where $\mid w(t) \mid$ is the number of temporal windows per video, 
$\mu_{\mathcal{P}_{w(t)}^c}$ represents the mean of the PSD response, and $\sigma_{\mathcal{P}_{w(t)}}$ denotes the standard deviation, while $k$ is an adjustable parameter.
We set $k = 3$ in our experiments.

The final predicted frequency over the 13 channels becomes: 
\begin{equation}
    f^* = \argmax_f ( \max_c \mathcal{S}^c (f) ).
\end{equation}

\newcommand{\red}[1]{\textcolor{red}{{#1}}}
\newcommand{\orange}[1]{\textcolor{orange}{{#1}}}
\newcommand{\green}[1]{\textcolor{green}{{#1}}}
\newcommand{\blue}[1]{\textcolor{blue}{{#1}}}
\newcommand{\purple}[1]{\textcolor{purple}{{#1}}}
\begin{figure}[ht!]
	\centering
    \fontsize{7}{7.5}\selectfont
	\begin{tabular}{c}
	    \includegraphics[width=0.9\textwidth]{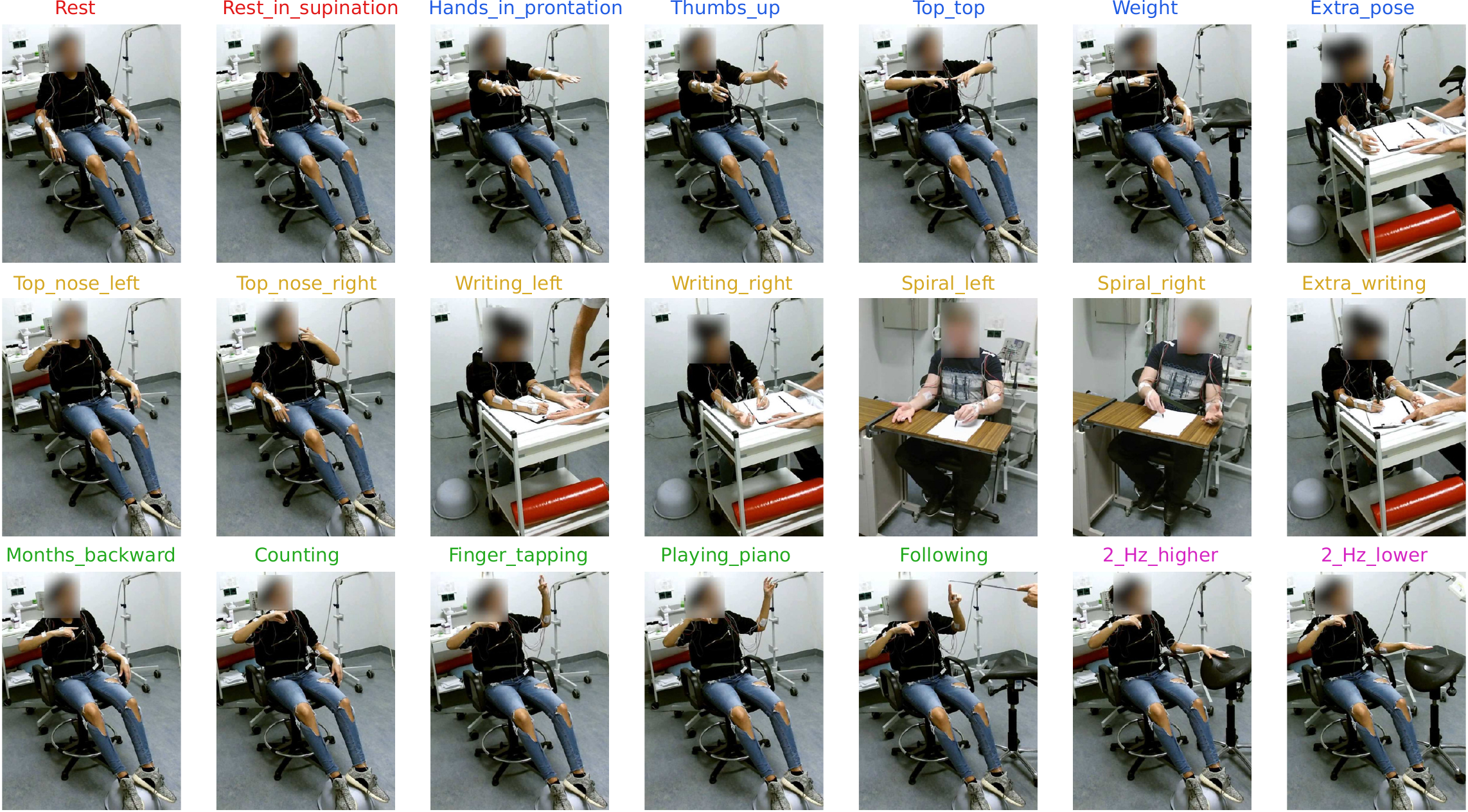} \\ 
	    (a) Recorded tasks.\\[5px]
        \begin{tabular}{l@{\hskip 0.1in}l}
            \toprule 
            Task & Description\\
            \midrule
            \multicolumn{2}{l}{\red{\textbf{Rest}}}\\
            Rest                        & Resting the arms on the chair handles.\\
            Rest\_in\_supination        & Resting the arms on the chair handles, hands in supination position.\\ \midrule
            \multicolumn{2}{l}{\blue{\textbf{Postures}}}\\
            Hands\_in\_pronation        & Both arms outstretched forward, hands in pronation position.\\
            Thumbs\_up                  & Both arms outstretched forward,  thumbs up.\\
            Top\_top                    & Both hands in front of the chest with tips of the index fingers almost touching\\
                                        & each other, elbows lifted sideways at approx. 90 degrees angle.\\
            Weight                      & Affected arm outstretched forward, with a weight (1 kg) attached to the wrist.\\
            Extra\_pose                 & Holding a pose proposed by the medical expert to better visualize the tremor.\\ \midrule
            \multicolumn{2}{l}{\orange{\textbf{Actions}}}\\
            Top\_nose\_left             & Touching the top of the nose with the left index finger.\\
            Top\_nose\_right            & Touching the top of the nose with the right index finger.\\
            Writing\_left               & Writing a given sentence with the left hand.\\
            Writing\_right              & Writing a given sentence with the right hand.\\
            Spiral\_left                & Drawing a spiral with the left hand.\\
            Spiral\_right               & Drawing a spiral with the right hand.\\
            Extra\_writing              & Extra writing task with a special pen, or diverging from the standard writing task \\
                                        & with the affected hand.\\ \midrule
            \multicolumn{2}{l}{\green{\textbf{Distraction}}}\\
            Months\_backward$^*$        & Naming the months backwards.\\
            Counting$^*$                & Counting backwards (100 minus 7).\\
            Finger\_tapping$^*$         & Tapping with the index finger and thumb of the contralateral hand.\\
            Playing\_piano$^*$          & Moving the thumb of the contralateral hand across all fingers, from the index \\
                                        & to the pinky finger and back. \\
            Following$^*$               & Following a moving pointer with the index finger of the contralateral hand.\\ \midrule
            \multicolumn{2}{l}{\purple{\textbf{Entrainment}}}\\
            2\_Hz\_higher$^*$           & Tapping with the contralateral hand in the rhythm of a flashing light, \\
                                        & 2 Hz higher than the estimated tremor frequency of the affected hand.\\
            2\_Hz\_lower$^*$            & Tapping with the contralateral hand in the rhythm of a flashing light, \\
                                        & 2 Hz lower than the estimated tremor frequency of the affected hand.\\ \midrule
            \multicolumn{2}{l}{$^*$During these tasks, the affected hand was kept in the posture in which the tremor was most pronounced }\\
            \multicolumn{2}{l}{ (i.e. arm on chair handle, arm outstretched with hand in pronation or thumbs up, or in front of the chest).}\\
            \bottomrule
	    \end{tabular}\\
	    (b) Explanation.\\
	\end{tabular}
    \caption{
	    (a) We record motor-disorder patients in 21 tasks. Each task may elicit a tremor.
	    (b) Short explanation of what each task involves.
    }
	\label{fig:tasks}
\end{figure}
\section{Experiments}
We test the considered frequency estimation approaches on our tremor patient dataset, \emph{TIM-Tremor}, 
containing a multitude of tasks.
The anonymized \emph{TIM-Tremor} patient data can be found at 
\href{https://doi.org/10.4121/uuid:522d14ed-3019-4206-b49e-a4e674b6440a}{https://doi.org/10.4121/uuid:522d14ed-3019-4206-b49e-a4e674b6440a}.
\subsection{Patient data evaluation}
\label{sec:patient}
\begin{figure}[t]
	\centering
	\begin{tabular}{c@{\hskip 0.3in}c@{\hskip 0.3in}c@{\hskip 0.3in}c}
	\includegraphics[width=0.16\textwidth]{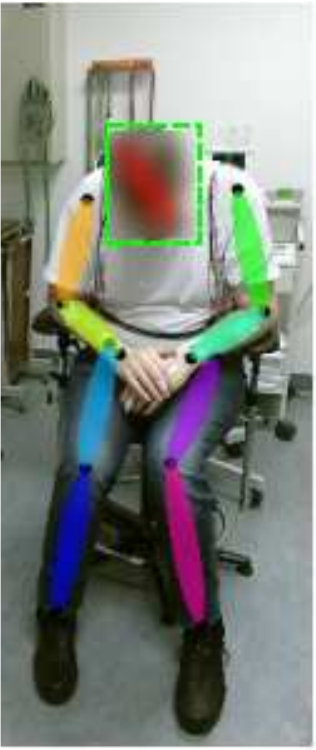} &
	\includegraphics[width=0.16\textwidth]{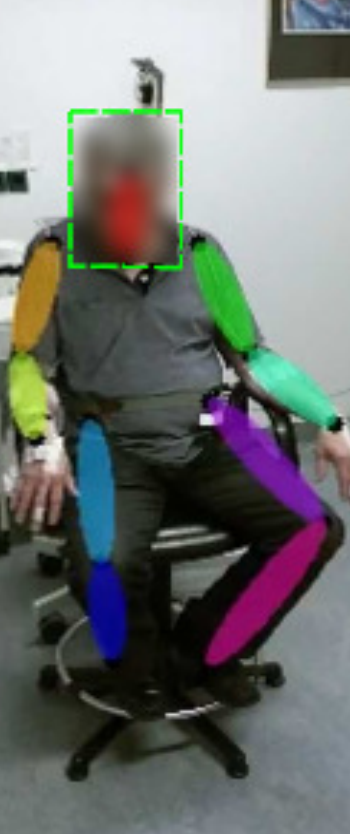} &
	\includegraphics[width=0.16\textwidth]{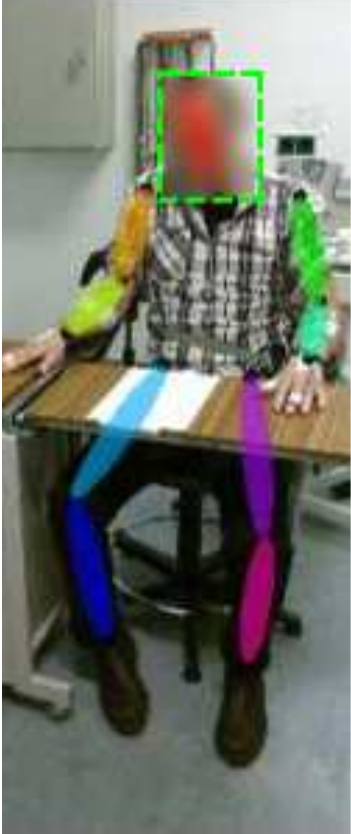} &
	\includegraphics[width=0.16\textwidth]{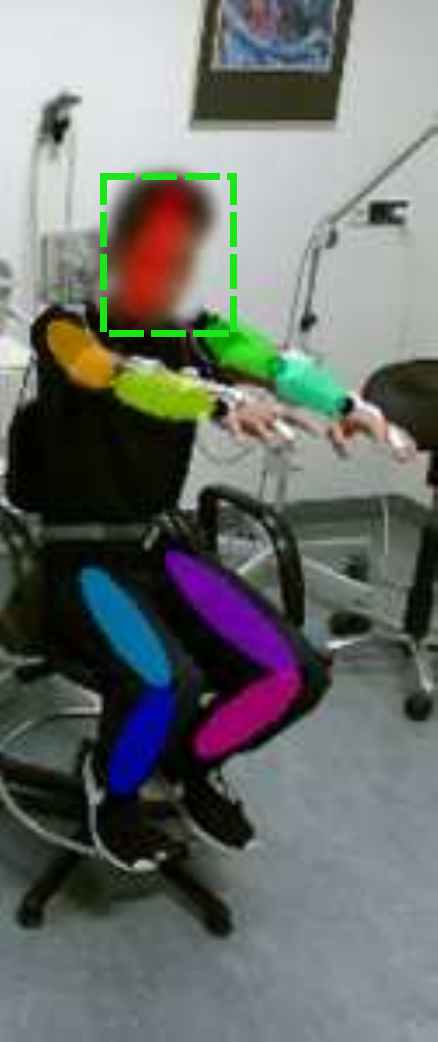} \\
	\end{tabular}
	\caption{Examples from the recording setup together with the predicted body joint locations using \cite{wei2016convolutional}. 
        We use this to obtain the location of the hand where we estimate the tremor frequency.
	}
	\label{fig:patient}
\end{figure}

\begin{figure}[t]
	\centering
	\begin{tabular}{c@{\hskip 0.2in}c}
	    \includegraphics[width=0.45\textwidth]{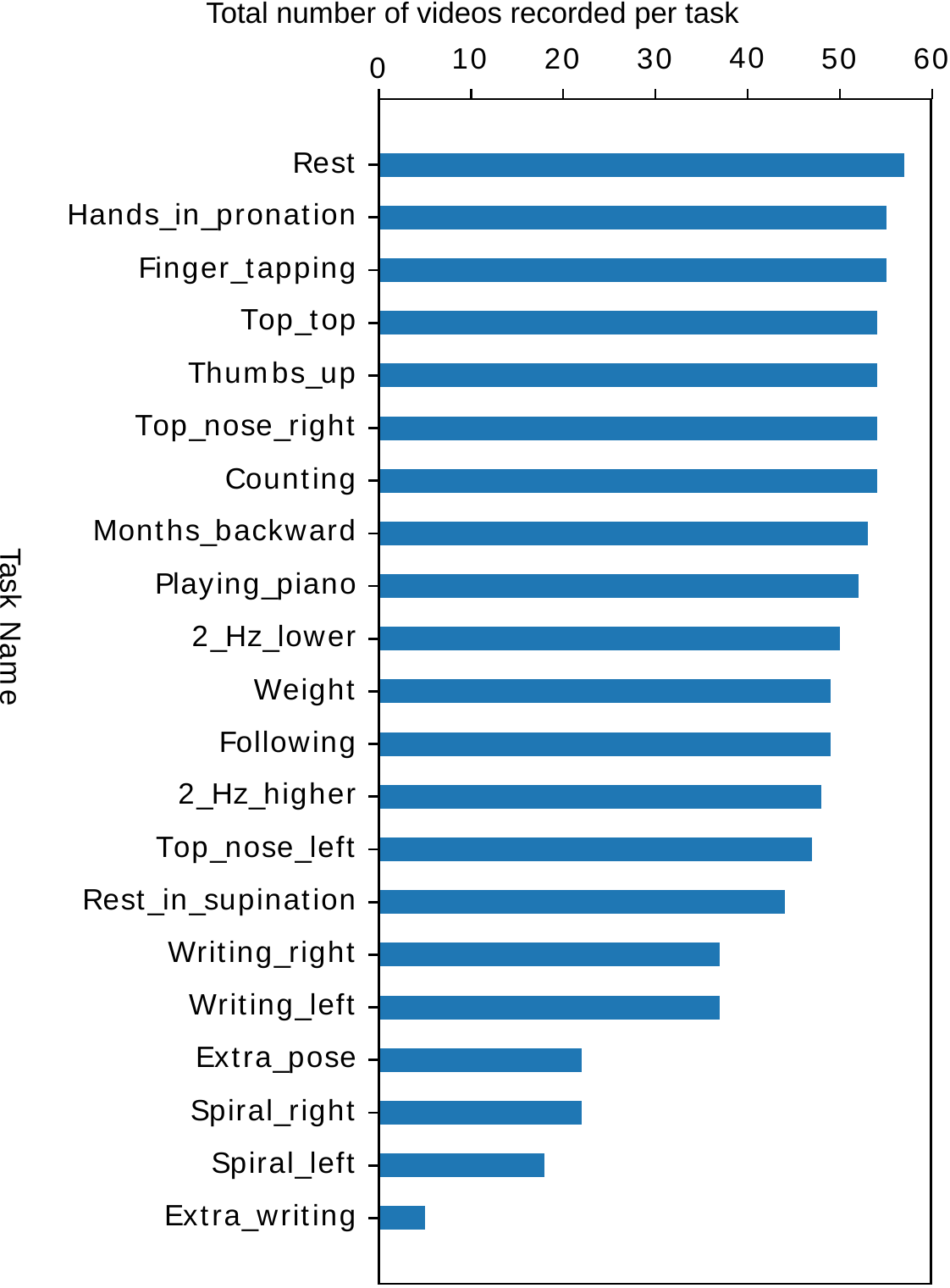} &
	    \includegraphics[width=0.45\textwidth]{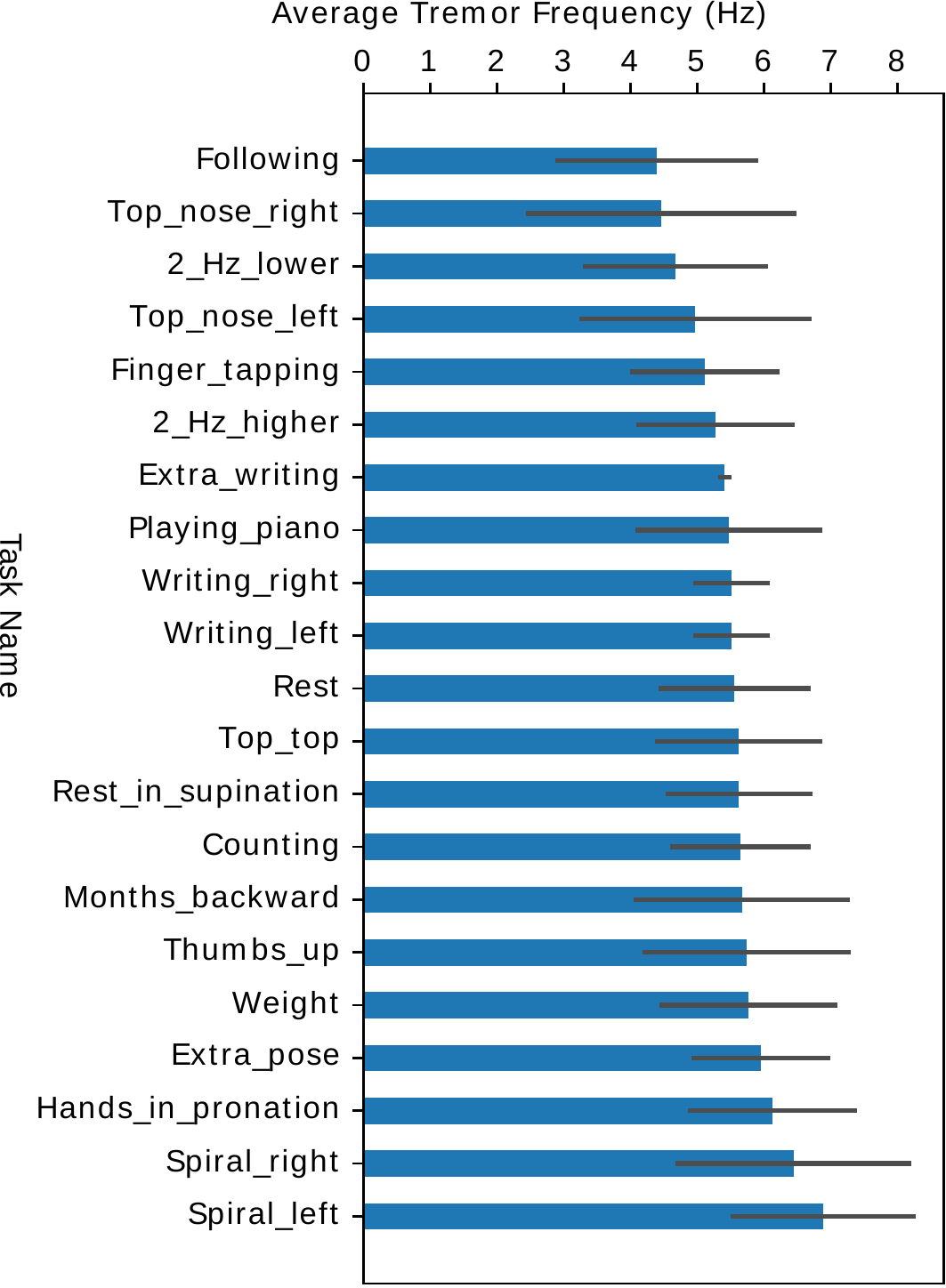} \\
        (a) & (b) \\
    \end{tabular}
	\caption{(a) Total number of videos recorded per task.
        (b) Average frequency and standard deviation for all tasks across all \records patient recordings. 
	}
	\label{fig:data-stat}
\end{figure}

\subsubsection{Data description.}
We recorded the \emph{TIM-Tremor} dataset, in which \records patients are videotaped sitting in a chair and performing a multitude of tasks.
The data is recorded with a Kinect$^{TM}$ v2 device, and it consists of short RGB videos of resolution $1920 \times 1080$ px, 
and associated depth video recordings of $512 \times 424$ px using a 16-bit encoding, as well as depth videos aligned with the RGB videos 
following the method in \cite{kooij2016sensecap}.
To reduce the storage requirements, we rescale the video resolution to $960 \times 540$ px.
The ground truth tremor frequency is measured on the wrist of the most affected hand: left\slash right. 
On this hand, during the recording, we position an accelerometer.
The accelerometer recordings are included in the dataset. 
The hand on which the accelerometer is positioned, is annotated in the dataset for each patient. 
Thus, for each patient and each performed task, we provide a set of recorded videos of approximately 1 minute each, together with a corresponding aligned depth map video, 
and the ground truth accelerometer recording from the most affected hand.
 
Data collection occurred in parallel to the standard tremor clinical evaluation. 
The standard tremor evaluation consists of a set of 21 tasks, which are illustrated in \fig{tasks}.(a) and described in \fig{tasks}.(b). 
The tasks vary with respect to the adopted posture: e.g. arm supported by the arm rest, or held outstretched in front of the patient, 
the amount of motion involved: e.g. rest -- no motion, or touching the top of the nose -- intention-oriented motion, 
as well as the focus of attention: e.g. distraction by mental task. 
Changes in tremor frequency between these tasks are analyzed by the medical expert to classify the tremor. 
For example, certain types of tremor are present across most or all tasks (e.g. ``Parkinsonian tremor"), 
while other types of tremor may only occur when performing a specific task (e.g. ``postural tremor" occurs only when a patient maintains a specific posture such as \emph{Thumbs\_up}), 
while other tremors may show considerable variation in tremor frequency between tasks (e.g. ``functional tremor").

\fig{patient} displays a few examples of the recording setup together with the estimated joint locations using \cite{wei2016convolutional}. 
In \fig{data-stat} we show the total number of videos recorded for each task, 
and the average hand tremor frequency, as estimated by the accelerometer,
together with the standard deviation, computed across all \records patients.
The average tremor frequency is around $5$ Hz, which is a common in tremor affections such as Parkinson and Dystonia.    

\subsubsection{Experimental evaluation.}
We estimate the body pose in the videos using the method in \cite{wei2016convolutional}, pretrained on the MPII dataset \cite{andriluka20142d}.
We apply the method a every frame.
We use a temporal window, $w(t)$, of 60 frames for frequency estimation.
Unless stated otherwise, we evaluate our method in terms of MAE (Mean Absolute Error) with respect to the ground truth frequency detected by the accelerometer.
We only evaluate on video segments in which a periodic tremor has been detected, using the accelerometer data.

\subsection{Exp. 1: Design choices}
In this experiment we test individual choices in the considered Lagrangian and Eulerian approaches. 
For the Lagrangian approach we test in \textbf{Exp 1.1} if removing the smooth trajectory, corresponding to the large motion of the hand,
helps the frequency estimation.
For the Eulerian approach, in \textbf{Exp 1.2} we test the added value of computing hand-tremor frequency over the phase information, 
rather than using only the intensity values of the image.

\begin{table}[t]
	\centering
	\begin{tabular}{l@{\hskip 0.5in}c@{\hskip 0.3in}c}
	\toprule
    Task                 & Lag\_no\_smooth (Hz) &   Lag\_with\_smooth (Hz) \\ \midrule
    2\_Hz\_higher        & 1.917 ($\pm$ 2.395)  &   \textbf{1.879} ($\pm$ 2.127) \\
    2\_Hz\_lower         & 2.248 ($\pm$ 2.770)  &   \textbf{1.721} ($\pm$ 2.266) \\ 
    Counting             & 1.731 ($\pm$ 2.336)  &   \textbf{1.377} ($\pm$ 2.246) \\
    Extra\_pose          & 3.590 ($\pm$ 2.369)  &   \textbf{1.918} ($\pm$ 1.328) \\ 
    Extra\_writing       & \textbf{1.968} ($\pm$ 0.000)  &   \textbf{1.968} ($\pm$ 0.000) \\
    Finger\_tapping      & 1.989 ($\pm$ 2.783)  &   \textbf{1.326} ($\pm$ 1.974) \\
    Following            & 1.607 ($\pm$ 1.745)  &   \textbf{1.312} ($\pm$ 1.728) \\
    Hands\_in\_pronation & 2.582 ($\pm$ 2.154)  &   \textbf{2.398} ($\pm$ 2.024) \\ 
    Months\_backward     & 2.544 ($\pm$ 2.703)  &   \textbf{2.031} ($\pm$ 2.500) \\
    Playing\_piano       & 2.443 ($\pm$ 2.826)  &   \textbf{2.033} ($\pm$ 2.516) \\
    Rest                 & \textbf{3.300} ($\pm$ 3.271)  &   3.395 ($\pm$ 3.226) \\
    Rest\_in\_supination & 2.889 ($\pm$ 3.228)  &   \textbf{2.059} ($\pm$ 2.248) \\
    Spiral\_left         & \textbf{6.721} ($\pm$ 1.896)  &   \textbf{6.721} ($\pm$ 1.896) \\
    Spiral\_right        & 3.246 ($\pm$ 1.762)  &   \textbf{3.148} ($\pm$ 1.803) \\
    Top\_nose\_left      & 3.743 ($\pm$ 3.262)  &   \textbf{3.688} ($\pm$ 3.242) \\
    Top\_nose\_right     & 1.928 ($\pm$ 2.323)  &   \textbf{1.771} ($\pm$ 2.204) \\
    Top\_top             & \textbf{1.388} ($\pm$ 1.797)  &   1.669 ($\pm$ 1.888) \\ 
    Thumbs\_up           & \textbf{1.694} ($\pm$ 1.807)  &   1.694 ($\pm$ 1.836) \\
    Weight               & \textbf{2.660} ($\pm$ 2.667)  &   2.795 ($\pm$ 2.569) \\ 
    Writing\_left        & \textbf{2.557} ($\pm$ 1.139)  &   \textbf{2.557} ($\pm$ 1.139) \\
    Writing\_right       & \textbf{2.557} ($\pm$ 1.139)  &   \textbf{2.557} ($\pm$ 1.139) \\ \midrule
    Average MAE          & 2.633 ($\pm$ 2.208)  &   \textbf{2.382} ($\pm$ 1.995) \\ 
	\bottomrule
	\end{tabular}
	\caption{\textbf{Exp 1.1:} MAE when comparing the Lagrangian method with trajectory smoothing by using the Kalman tracker --- \emph{Lag\_with\_smooth}, 
        versus not using trajectory smoothing, \emph{Lag\_no\_smooth}. 
        \emph{Lag\_with\_smooth} performs slightly better than the default Lagrangian method, \emph{Lag\_no\_smooth}.
        We highlight in bold the better performing method (lower is better). 
	}
	\label{tab:lag}
\end{table}
\bigskip\noindent \textbf{Exp 1.1: The need of trajectory smoothing.}
We experimentally compare two variants of the Lagrangian frequency estimation. 
The \emph{Lag\_no\_smooth} variant uses raw hand trajectory points as computed by the pose estimation algorithm. 
The \emph{Lag\_with\_smooth} variant removes the large motion of the hand obtained by subtracting the output of a Kalman tracker, 
which in effect retains only the small motions. 
The MAE numbers in \tab{lag} show that removing the large motion by using the Kalman tracker is beneficial to the overall performance.
This is explained by the fact that subtracting the trajectory returned by the Kalman tracker from the original hand trajectory works as a data detrending step.
This allows for the frequency to be estimated only over the small tremor motion.
 
\begin{table}[t]
	\centering
	\begin{tabular}{l@{\hskip 0.5in}c@{\hskip 0.3in}c}
	\toprule
    Task                    & Euler\_gray (Hz)      &   Euler\_phase (Hz) \\ \midrule
    2\_Hz\_higher           & 0.882 ($\pm$ 1.142)   &   \textbf{0.857} ($\pm$ 1.533) \\
    2\_Hz\_lower            & 1.335 ($\pm$ 2.022)   &   \textbf{0.984} ($\pm$ 1.333) \\
    Counting                & 0.767 ($\pm$ 1.252)   &   \textbf{0.472} ($\pm$ 0.780) \\
    Extra\_pose             & \textbf{1.180} ($\pm$ 2.006)   &   1.623 ($\pm$ 1.185) \\
    Extra\_writing          & \textbf{0.984} ($\pm$ 0.000)   &   \textbf{0.984} ($\pm$ 0.000) \\
    Finger\_tapping         & 0.492 ($\pm$ 0.893)   &   \textbf{0.385} ($\pm$ 0.647) \\
    Following               & 0.820 ($\pm$ 1.327)   &   \textbf{0.557} ($\pm$ 0.503) \\
    Hands\_in\_pronation    & 1.271 ($\pm$ 1.755)   &   \textbf{1.066} ($\pm$ 1.506) \\
    Months\_backward        & 1.133 ($\pm$ 1.848)   &   \textbf{1.219} ($\pm$ 1.933) \\
    Playing\_piano          & 1.148 ($\pm$ 1.832)   &   \textbf{1.148} ($\pm$ 1.714) \\
    Rest                    & 1.459 ($\pm$ 1.759)   &   \textbf{1.253} ($\pm$ 1.770) \\
    Rest\_in\_supination    & \textbf{1.475} ($\pm$ 1.919)   &   1.537 ($\pm$ 1.728) \\
    Spiral\_left            & 3.278 ($\pm$ 1.671)   &   \textbf{2.951} ($\pm$ 1.391) \\
    Spiral\_right           & 3.246 ($\pm$ 2.936)   &   \textbf{2.509} ($\pm$ 2.021) \\
    Top\_nose\_left         & 2.595 ($\pm$ 2.216)   &   \textbf{1.776} ($\pm$ 2.008) \\
    Top\_nose\_right        & 3.108 ($\pm$ 2.739)   &   \textbf{2.164} ($\pm$ 2.015) \\
    Top\_top                & 0.860 ($\pm$ 1.311)   &   \textbf{0.720} ($\pm$ 1.200) \\
    Thumbs\_up              & 1.002 ($\pm$ 1.419)   &   \textbf{1.002} ($\pm$ 1.273) \\
    Weight                  & 1.207 ($\pm$ 1.695)   &   \textbf{0.961} ($\pm$ 1.226) \\
    Writing\_left           & \textbf{0.394} ($\pm$ 0.573)   &   0.492 ($\pm$ 0.538) \\
    Writing\_right          & \textbf{0.394} ($\pm$ 0.573)   &   0.492 ($\pm$ 0.538) \\ \midrule
    Average MAE             & 1.382 ($\pm$ 1.566)   &   \textbf{1.198} ($\pm$ 1.278) \\
    \bottomrule
	\end{tabular}
	\caption{
        \textbf{Exp 1.2:} MAE showing the added value of the phase information.
        We compare the \emph{Euler\_gray} --- Eulerian frequency estimation over grayscale hand-images, with 
        \emph{Euler\_phase} -- Eulerian frequency estimation over 12 phase-images and 1 grayscale image. 
        Adding the 12 extra phase-images is beneficial for the frequency estimation. 
        We highlight in bold the better performing method (lower is better).
	}
	\label{tab:eulerian}
\end{table}
\bigskip\noindent \textbf{Exp 1.2: The added value of using phase-images.}
For both considered Eulerian approaches we stabilize the trajectory along which we measure the tremor by using the Kalman tracker, and subsequently 
perform the frequency estimation over the complete hand window. 
In \tab{eulerian} we test the added value of using phase information for frequency estimation.
We compare two variants. The \emph{Euler\_gray} variant estimates the frequency over gray-scale pixels over  gray-scale hand-images, obtained by cropping the hand location 
along the smoothed trajectory of the hand. The \emph{Euler\_phase} variant adds the 12 phase channels as detailed in section~\ref{ssec:eul_method}.    
The phase channels allow the \emph{Euler\_phase} to more precisely capture the small motion corresponding to the tremor, because the phase is effective for describing motion. 
The MAE numbers in \tab{eulerian} validate that adding the phase information is beneficial for the hand-tremor frequency estimation.  

\subsection{Exp 2: Eulerian versus Lagrangian tremor frequency estimation}
\begin{figure}[t]
	\centering
	\includegraphics[width=1.0\textwidth]{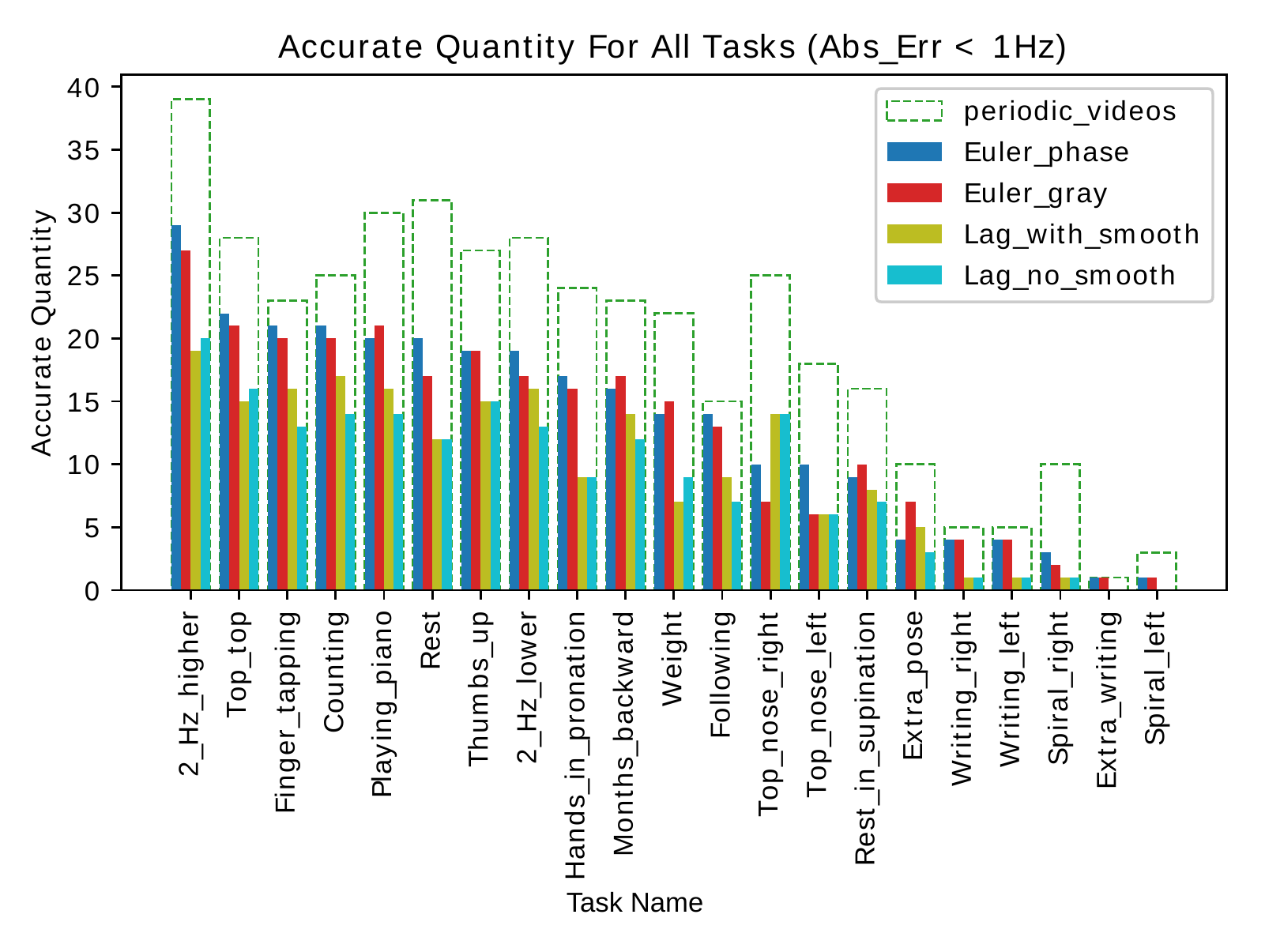} 
    \caption{\textbf{Exp 2:} We report accuracy on all recorded tasks, over the \records patient recordings (higher is better).  
            We consider the hand-tremor frequency to be correctly estimated for a task if the MAE (Mean Absolute Error) for that task is lower than 1 Hz. 
            We plot in dotted green line the total number of videos recorded for each task, on which we have detected a periodic tremor.
            For each of our considered methods we show the number of videos for which the frequency was correctly estimated. 
            On average the Eulerian methods perform better than the Lagrangian methods.}
    \label{fig:accuracy}
\end{figure}

In \fig{accuracy} we display the accuracy of our proposed frequency estimation methods over the complete set of \records patient recordings, for all tasks.
We show in dotted green line the number of videos per task where a periodic tremor was detected, according to the accelerometer data.
In corresponding color, we show the number of videos in which we have correctly estimated the hand-tremor frequency, for each frequency estimation method:
\emph{Euler\_phase} is the Eulerian method using 12 phase-channels and 1 grayscale channel; \emph{Euler\_gray} is the Eulerian method on image intensity information only; 
\emph{Lag\_no\_smooth} is the Lagrangian method without Kalman trajectory smoothing; \emph{Lag\_with\_smooth} is the Lagrangian method with Kalman trajectory smoothing. 
We consider an estimated tremor frequency to be correct if the MAE between the accelerometer frequency and the one estimated by the method is lower than 1~Hz. 

\begin{figure}[t]
	\centering
	\begin{tabular}{c@{\hskip 0.3in}c}
	\includegraphics[width=0.4\textwidth]{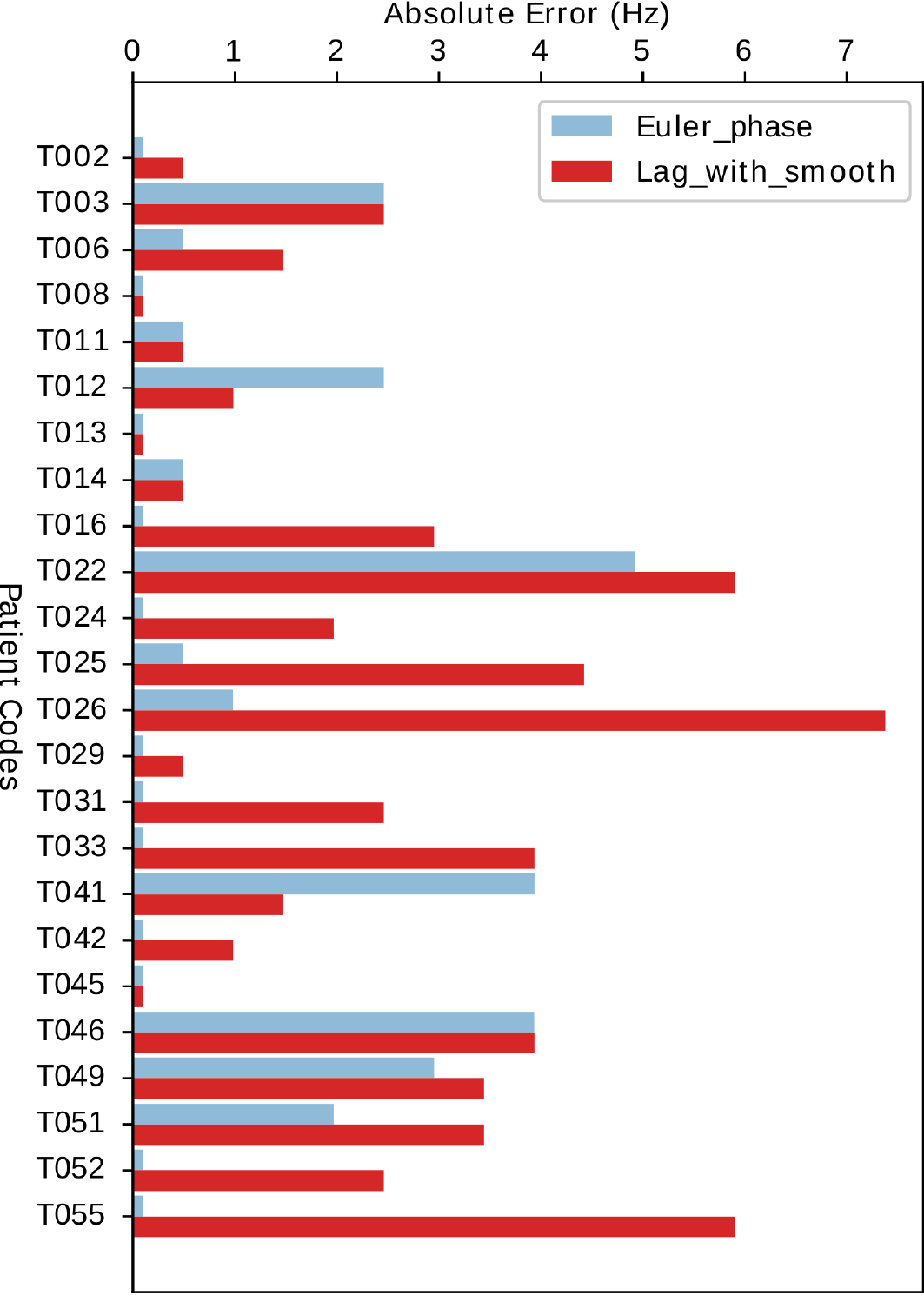} &
	\includegraphics[width=0.40\textwidth]{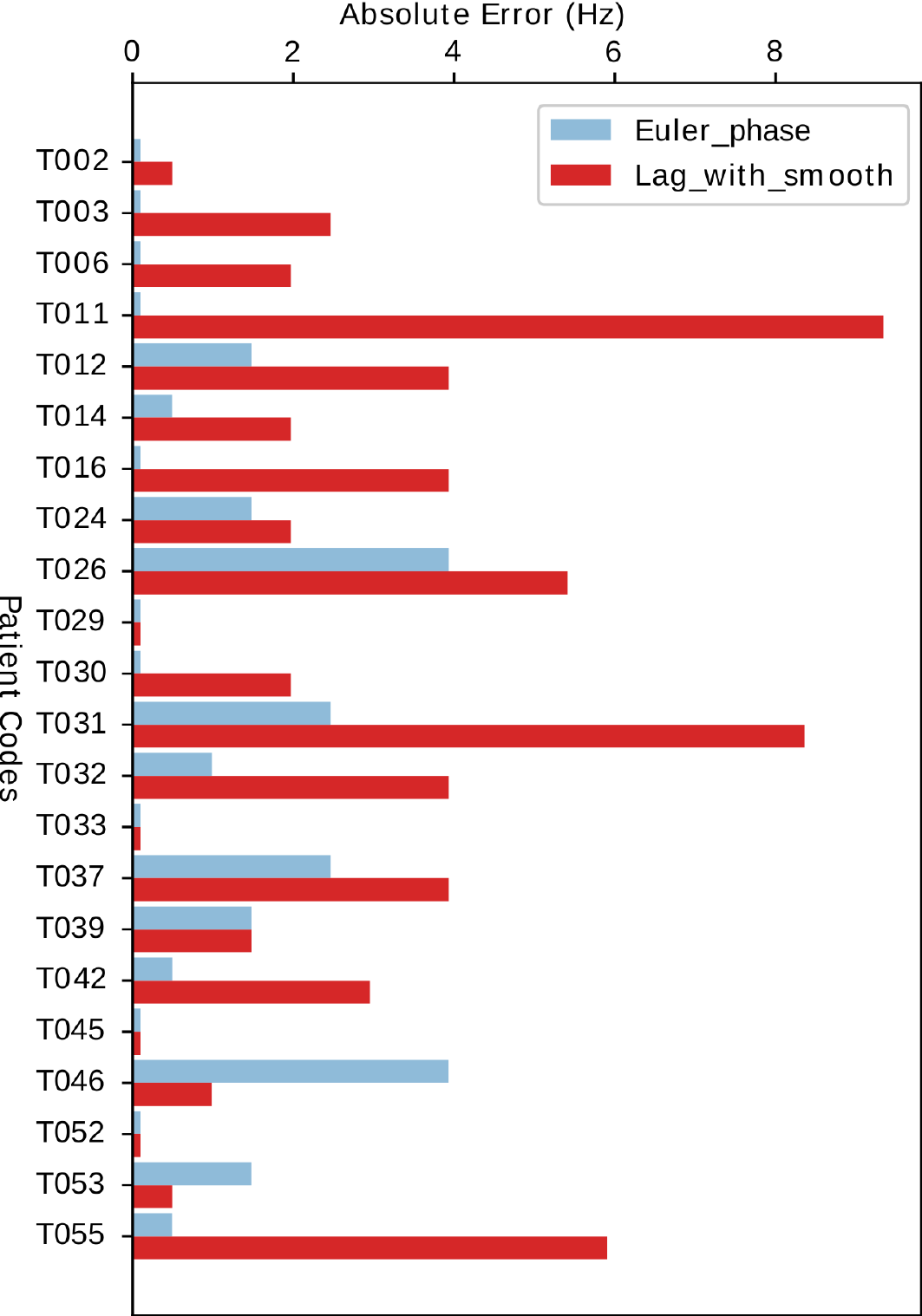} \\
    (a) \emph{Hands\_in\_prontation} task. & (b) \emph{Weight} task.
	\end{tabular}
	\caption{ 
       \textbf{Exp 2:} 
        The MAE per patient, for the two tasks where the Eulerian methods performed better than the Lagrangian methods (lower is better). 
        To avoid over-cluttering the image, we plot only the best performing Lagrangian method: \emph{Lag\_with\_smooth} --- Lagrangian method with Kalman trajectory smoothing, 
        and the best performing Eulerian method: \emph{Euler\_phase} --- Eulerian method using 12 phase channels and 1 grayscale channel.
        The Lagrangian method makes large frequency estimation mistakes on a few patient videos, while the Eulerian method is more precise on some of the patient videos.        
        (Note: for certain patients the task has not been recorded or no stable frequency, according to the accelerometer, has been found.)
	}
	\label{fig:expl_res}
\end{figure}

\fig{accuracy} shows that on average the Eulerian frequency estimation methods are more precise than the Lagrangian methods.
The gain of using the Eulerian approaches is especially clear for the \emph{Weight} task and the \emph{Hands\_in\_pronation} task. 
\fig{expl_res} displays the MAE scores per patient for these two tasks.
To avoid over-cluttering the image, we only show the best Lagrangian method: \emph{Lag\_with\_smooth}, Lagrangian with Kalman trajectory smoothing, and 
the best Eulerian method: \emph{Euler\_phase}, Eulerian over 12 phase channels and 1 grayscale channel. 
The Eulerian method gives more precise frequency estimates for some of the patient recordings, while for others it performs similar to the Lagrangian method.
The tasks are not characterized by large hand motion. 
The gain of the Eulerian method over the Lagrangian is explained by the Eulerian method better describing the subtle changes in image information over time at the hand location. 
Therefore, the Eulerian method more accurately captures the tremor in tasks that do not involve large hand motion, but exhibit small motion.

\section{Conclusions}
We consider the task of hand-tremor frequency estimation from RGB videos. 
We propose two different approaches for measuring human hand-tremor frequencies: 
(a) Lagrangian hand-tremor frequency estimation, using the trajectory of the hand motion in the image plane throughout the video, to assess the hand-tremor frequency; and
(b) Eulerian hand-tremor frequency estimation, which measures the change in the image information over time, at the location of hand in the image plane. 
We experimentally evaluate two variants of each approach on our proposed \emph{TIM-Tremor} dataset containing \records patient recordings performing a multitude of tasks.
From our experimental analysis we learned that the Eulerian approaches are more accurate on average than the Lagrangian methods, with the difference being substantial on tasks on 
which there is a limited amount of large hand motion, but where there is a small hand-tremor motion present. 

\bigskip\noindent\textbf{Acknowledgements.}
{\small 
This work is part of the research  program Technology in Motion (TIM) (628.004.001), which is
financed by the Netherlands Organisation for Scientific Research (NWO).
Many thanks for the help with data collection to Elma Ouwehand, MSc.}

{\small
\bibliographystyle{splncs04}
\bibliography{main}
}

\end{document}